%% file: ms.tex
\documentclass{article} 
\usepackage{iclr2020_conference,times}
\iclrfinalcopy

\usepackage[utf8]{inputenc} 
\usepackage[T1]{fontenc}    
\usepackage{hyperref}       
\usepackage{url}            
\usepackage{booktabs}       
\usepackage{amsfonts}       
\usepackage{nicefrac}       
\usepackage{microtype}      
\usepackage{algorithm}
\usepackage{algorithmic}

\usepackage{float}
\usepackage{standalone}
\usepackage{graphicx}
\usepackage{caption} 
\usepackage{subcaption}
\usepackage{wrapfig}

\usepackage{tikz}
\usetikzlibrary{
  arrows.meta,shapes.geometric,decorations.pathmorphing,
  backgrounds,positioning,fit,petri
}

\usepackage{color}
\usepackage{setspace}
\usepackage{bm,amsthm}
\usepackage{mathtools}
\usepackage[inline]{enumitem}
\input{math_cmds}
\renewcommand\cite{\citep}%

\newcommand*{\affaddr}[1]{#1}
\newcommand*{\affmark}[1][*]{\textsuperscript{\textnormal{#1}}}
\newcommand*{\email}[1]{\texttt{#1}}

\title{Relational State-Space Model \\ for Stochastic Multi-Object Systems}

\author{%
  Fan Yang\affmark[\dag], Ling Chen\thanks{Corresponding author.}\affmark[\;\;\dag], Fan Zhou\affmark[\dag], Yusong Gao\affmark[\ddag], Wei Cao\affmark[\ddag] \\
  \affaddr{\affmark[\dag]College of Computer Science and Technology, Zhejiang University, Hangzhou, China} \\
  \affaddr{\affmark[\ddag]Alibaba Group, Hangzhou, China} \\
  \email{\{fanyang01,lingchen,fanzhou\}@zju.edu.cn} \\
  \email{\{jianchuan.gys,mingsong.cw\}@alibaba-inc.com} \\
}

\begin{document}

\maketitle

\input{tex/abstract}

\input{tex/introduction}
\input{tex/background}
\input{tex/r-ssm}
\input{tex/related}
\input{tex/exp}
\input{tex/conclusion}
\input{tex/acknowledgement}

\bibliography{bib/deep-ssm,bib/gnn,bib/multi-agent,bib/multi-agent-latent,bib/flow,bib/inference,bib/misc}
\bibliographystyle{iclr2020_conference}

\newpage

\appendix
\input{tex/appendix}

\end{document}

%% file: math_cmds.tex
\newcommand*{\V}[1]{\mathbf{#1}}
\newcommand*{\M}[1]{\mathbf{#1}}
\newcommand*{\BR}{\mathbb{R}}

\newcommand*{\BE}{\mathbb{E}}
\newcommand*{\CG}{\mathcal{G}}
\newcommand*{\CV}{\mathcal{V}}
\newcommand*{\CE}{\mathcal{E}}
\newcommand*{\CA}{\mathcal{A}}
\newcommand*{\CM}{\mathcal{M}}
\newcommand*{\CN}{\mathcal{N}}
\newcommand*{\CL}{\mathcal{L}}

\newcommand*{\CX}{\mathcal{X}}
\newcommand*{\CY}{\mathcal{Y}}
\newcommand*{\CZ}{\mathcal{Z}}

\newcommand*{\MH}{\mathbf{H}}
\newcommand*{\MZ}{\mathbf{Z}}
\newcommand*{\MX}{\mathbf{X}}
\newcommand*{\MW}{\mathbf{W}}
\newcommand*{\MB}{\mathbf{B}}
\newcommand*{\MU}{\mathbf{U}}
\newcommand*{\MV}{\mathbf{V}}

\newcommand*{\Va}{\mathbf{a}}
\newcommand*{\Vb}{\mathbf{b}}
\newcommand*{\Vc}{\mathbf{c}}
\newcommand*{\Ve}{\mathbf{e}}
\newcommand*{\Vg}{\mathbf{g}}
\newcommand*{\Vh}{\mathbf{h}}

\newcommand*{\Vv}{\mathbf{v}}
\newcommand*{\Vu}{\mathbf{u}}
\newcommand*{\Vx}{\mathbf{x}}
\newcommand*{\Vz}{\mathbf{z}}

\DeclareMathAlphabet{\mathsfit}{\encodingdefault}{\sfdefault}{m}{sl}
\SetMathAlphabet{\mathsfit}{bold}{\encodingdefault}{\sfdefault}{bx}{n}
\newcommand{\tens}[1]{\bm{\mathsfit{#1}}}
\def\TE{{\tens{E}}}
\def\TX{{\tens{X}}}

\newcommand*{\BF}[1]{\textbf{#1}}

%% file: tex/abstract.tex
\begin{abstract}

Real-world dynamical systems often consist of multiple stochastic subsystems
that interact with each other. Modeling and forecasting the behavior of such
dynamics are generally not easy, due to the inherent hardness in understanding
the complicated interactions and evolutions of their constituents. This paper
introduces the relational state-space model (R-SSM), a sequential hierarchical
latent variable model that makes use of graph neural networks (GNNs) to
simulate the joint state transitions of multiple correlated objects. By
letting GNNs cooperate with SSM, R-SSM provides a flexible way to incorporate
relational information into the modeling of multi-object dynamics. We further
suggest augmenting the model with normalizing flows instantiated for
vertex-indexed random variables and propose two auxiliary contrastive
objectives to facilitate the learning. The utility of R-SSM is empirically
evaluated on synthetic and real time series datasets.

\end{abstract}

%% file: tex/introduction.tex
\section{Introduction}

Many real-world dynamical systems can be decomposed into smaller interacting
subsystems if we take a fine-grained view. For example, the trajectories of
coupled particles are co-determined by per-particle physical properties (e.g.,
mass and velocity) and their physical interactions (e.g., gravity); traffic
flow can be viewed as the coevolution of a large number of vehicle dynamics.
Models that are able to better capture the complex behavior of such
multi-object systems are of wide interest to various communities, e.g.,
physics, ecology, biology, geoscience, and finance.

State-space models (SSMs) are a wide class of sequential latent variable
models (LVMs) that serve as workhorses for the analysis of dynamical systems
and sequence data. Although SSMs are traditionally designed under the guidance
of domain-specific knowledge or tractability consideration, recently
introduced deep SSMs \cite{deep-lvm} use neural networks (NNs) to parameterize
flexible state transitions and emissions, achieving much higher expressivity.
To develop deep SSMs for multi-object systems, graph neural networks (GNNs)
emerge to be a promising choice, as they have been shown to be fundamental NN
building blocks that can impose \emph{relational inductive bias} explicitly
and model complex interactions effectively \cite{gn}.

Recent works that advocate GNNs for modeling multi-object dynamics mostly make
use of GNNs in an autoregressive (AR) fashion. AR models based on recurrent
(G)NNs can be viewed as special instantiations of SSMs in which the state
transitions are restricted to being deterministic \cite[Section 4.2]{deep-lvm}.
Despite their simplicity, it has been pointed out that their modeling
capability is bottlenecked by the deterministic state transitions
\cite{vrnn,srnn} and the oversimplified observation distributions
\cite{softmax}.

In this study, we make the following contributions:
\begin{enumerate*}[label={(\roman*)},font={\bfseries}]
	\item We propose the relational state-space model (R-SSM), a novel
		hierarchical deep SSM that simulates the stochastic state transitions of
		interacting objects with GNNs, extending GNN-based dynamics
		modeling to challenging stochastic multi-object systems.
	\item We suggest using the graph normalizing flow (GNF) to construct
		expressive joint state distributions for R-SSM, further enhancing its
		ability to capture the joint evolutions of correlated stochastic
		subsystems.
	\item We develop structured posterior approximation to learn R-SSM using
		variational inference and introduce two auxiliary training objectives
		to facilitate the learning.
\end{enumerate*}

Our experiments on synthetic and real-world time series datasets show that
R-SSM achieves competitive test likelihood and good prediction performance
in comparison to GNN-based AR models and other sequential LVMs.
The remainder of this paper is organized as follows: Section 2 briefly reviews
neccesary preliminaries. Section 3 introduces R-SSM formally and presents the methods to learn R-SSM from observations. Related work is summarized in Section 4 and experimental evaluation is presented in Section 5. We conclude the paper in Section 6.

%% file: tex/background.tex
\section{Preliminaries}

In this work, an attributed directed graph is given by a 4-tuple: $\CG =
(\CV, \CE, \MV, \TE)$, where $\CV = [N] \coloneqq \{1, \dotsc, N\}$ is the set
of vertices, $\CE \subseteq [N] \times [N]$ is the set of edges, $\MV \in
\BR^{N \times d_v}$ is a matrix of static vertex attributes, and
$\TE \in \BR^{N \times N \times d_e}$ is a sparse tensor storing the static
edge attributes. The set of
direct predecessors of vertex $i$ is notated as $\CN^{-}_i = \left\{ p \middle| (p, i) \in \CE \right\}$. We use the notation $\Vx_i$ to refer to the $i$-th row of matrix $\MX$ and write $\Vx_{ij}$ to indicate the $(i, j)$-th
entry of tensor $\TX$ (if the corresponding matrix or tensor appears in the context). For sequences, we write $\Vx_{\leq t} = \Vx_{1:t} \coloneqq (\Vx_1, \ldots, \Vx_t)$ and switch to $\Vx_t^{(i)}$ for referring to the $i$-th row of matrix $\MX_t$.

\subsection{Graph Neural Networks}

GNNs are a class of neural networks developed to process graph-structured data
and support relational reasoning. Here we focus on vertex-centric GNNs that
iteratively update the vertex representations of a graph $\CG$ while being
\emph{equivariant} \cite{eq-gn} under vertex relabeling.
Let $\MH \in \BR^{N \times d}$ be a
matrix of vertex representations, in which the $i$-th row $\Vh_i \in \BR^d$ is
the vectorized representation attached to vertex $i$. Conditioning on the
static graph structure and attributes given by $\CG$, a GNN just takes the
vertex representations $\MH$ along with some graph-level context $\Vg \in
\BR^{d_g}$ as input and returns new vertex representations $\MH'\in \BR^{N
\times d'}$ as output, i.e., $\MH' = \text{GNN}(\CG, \Vg, \MH)$.

When updating the representation of vertex $i$ from $\Vh_i$ to
$\Vh_i^{\prime}$, a GNN takes the representations of other nearby vertices
into consideration. Popular GNN variants achieve this through a multi-round message
passing paradigm, in which the vertices repeatedly send messages to their neighbors,
aggregate the messeages they received, and update their own representations
accordingly. Formally, the operations performed by a basic block of a
message-passing GNN are defined as follows:
\begin{align}
\label{eq:gnn-msg-fn}
\forall (j, i) \in \CE &: &
\CM_{j \to i} &= \text{MESSAGE}\left( \Vg, \Vv_j, \Vv_i, \Ve_{ji}, \Vh_j, \Vh_i \right)
\\
\label{eq:gnn-agg-fn}
\forall i \in \CV &: &
\CA_i &= \text{AGGREGATE}\left( \left\{ \CM_{p \to i} \right\}_{p \in \CN^{-}_i} \right)
\\
\label{eq:gnn-cmb-fn}
\forall i \in \CV &: &
\Vh'_i &= \text{COMBINE}\left( \Vg, \Vv_i, \Vh_i, \CA_i \right)
\end{align}
Throughout this work, we implement Equations \eqref{eq:gnn-msg-fn} and \eqref{eq:gnn-agg-fn} by adopting a multi-head attention mechanism similar to \citet{transformer} and
\citet{gat}. For Equation \eqref{eq:gnn-cmb-fn}, we use either a RNN cell or a residual block \cite{he}, depending on whether the inputs to GNN are RNN states or not. We write such a block as $\MH' = \text{MHA}( \CG, \Vg, \MH )$ and give its detailed implementation in the Appendix. A GNN simply stacks $L$ separately-parameterized MHA blocks and iteratively computes
$\MH \eqqcolon \MH^{(0)}, \ldots, \MH^{(L)} \eqqcolon \MH'$, in which $\MH^{(l)} = \text{MHA}( \CG, \Vg, \MH^{(l-1)} )$ for $l = 1, \ldots, L$. We write this construction as $\MH' = \text{GNN}(\CG, \Vg, \MH)$ and treat it as a black box to avoid notational clutter.

\subsection{State-Space Models}

State-space models are widely applied to analyze dynamical systems whose true
states are not directly observable. Formally, an SSM assumes the dynamical
system follows a latent state process $\left\{\Vz_t \right\}_{t \geq 1}$,
which possibly depends on exogenous inputs $\{\Vu_t \}_{t \geq 1}$.
Parameterized by some (unknown) static parameter $\theta$, the
latent state process is characterized by an initial density $\Vz_1 \sim
\pi_{\theta}(\cdot | \Vu_1)$ and a transition density $\Vz_{t+1} \sim f_{\theta}\left(
\cdot \middle| \Vz_{\leq t}, \Vu_{\leq t+1} \right)$. Moreover, at each time
step, some noisy measurements of the latent state are observed through an
observation density: $\Vx_t \sim g_{\theta}\left( \cdot \middle| \Vz_{\leq t}, \Vu_{\leq t}
\right)$ . The joint density of $\Vx_{1:T}$ and $\Vz_{1:T}$ factors as:
$
p(\Vx_{1:T}, \Vz_{1:T} | \Vu_{1:T}) =  \pi_{\theta}( \Vz_1 | \Vu_1 )
\prod_{t=2}^{T}{f_{\theta}( \Vz_t | \Vz_{<t}, \Vu_{\leq t} )}
\prod_{t=1}^{T}{g_{\theta}( \Vx_t | \Vz_{\leq t}, \Vu_{\leq t} )}
$.

The superior expressiveness of SSMs can be seen from the fact that the
marginal predictive distribution $p(\Vx_t | \Vx_{<t}, \Vu_{\leq t}) = \int
p(\Vx_t, \Vz_{\leq t} | \Vx_{<t}, \Vu_{\leq t}) \: \mathrm{d}\Vz_{\leq t}$ can
be far more complex than unimodal distributions and their finite mixtures that
are common in AR models. Recently developed deep SSMs use RNNs to compress
$\Vz_{\leq t}$ (and $\Vu_{\leq t}$) into fixed-size vectors to achieve
tractability. As shown in next section, R-SSM can be viewed as
enabling multiple individual deep SSMs to communicate.

\subsection{Normalizing Flows}

Normalizing flows \cite{nf} are invertible transformations that have the capability to
transform a simple probability density into a complex one (or vice versa). Given two domains $\CX \subseteq \BR^D$ and $\CY \subseteq \BR^D$, let
$f: \CX \to \CY$ be an invertible mapping with inverse $f^{-1}$. Applying
$f$ to a random variable $\Vz \in \CX$ with density $p(\Vz)$, by the \emph{change
of variables} rule, the resulting random variable $\Vz' = f(\Vz) \in \CY$ will have a
density:
\[
p(\Vz') = p(\Vz) \left| \det \frac{\partial f^{-1}}{\partial \Vz'} \right|
= p(\Vz) \left| \det \frac{\partial f}{\partial \Vz} \right|^{-1}
\]
A series of invertible mappings with cheap-to-evaluate determinants can be chained together to achieve complex transformations while retaining efficient density calculation. This provides a powerful way to construct expressive distributions.

%% file: tex/r-ssm.tex
\section{Relational State-Space Model}

Suppose there is a dynamical system that consists of multiple interacting
objects, and observing this system at a specific time is accomplished by
acquiring measurements from every individual object simultaneously. We further assume these objects are homogeneous, i.e., they share the same
measurement model, and leave systems whose constituents are nonhomogeneous for
future work. To generatively model a time-ordered series of observations collected
from this system, the straightforward approach that builds an individual SSM for
each object is usually unsatisfactory, as it simply assumes the state of each
object evolves independently and ignores the interactions between objects.
To break such an independence assumption, our main idea is to let
multiple individual SSMs interact through GNNs, which are
expected to capture the joint state transitions of correlated objects well.

\begin{figure}
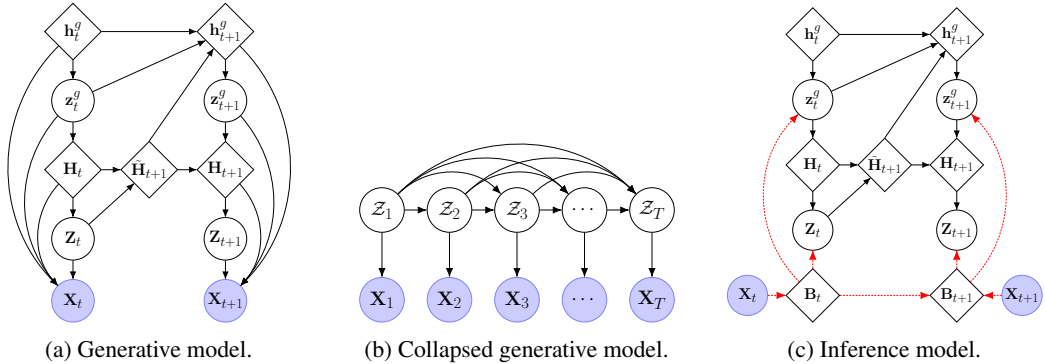

  \label{fig:gen-model}
  \centering
  \begin{subfigure}[b]{0.3\textwidth}
      \includestandalone[width=\textwidth]{img/generative-model}
      \caption{Generative model.}
      \label{fig:gen-model}
  \end{subfigure}
  \hfill
  \begin{subfigure}[b]{0.3\textwidth}
  	  \includestandalone[width=\textwidth]{img/generative-model-collapsed}
  	  \caption{Collapsed generative model.}
  	  \label{fig:gen-model-collapsed}
  \end{subfigure}
  \hfill
  \begin{subfigure}[b]{0.3\textwidth}
      \includestandalone[width=\textwidth]{img/proposal}
      \caption{Inference model.}
      \label{fig:proposal}
  \end{subfigure}
  \caption{Graphical structures of R-SSM. Diamonds represent deterministic states and circles represent random variables. To be concise, the dependencies on the graph $\CG$ and exogenous inputs $\MU_{1:T}$ are not shown. (b) is the result of collapsing all deterministic states in (a) and writing $\CZ_t = (\Vz_t^g, \MZ_t)$. In (c), solid lines represent the computation shared with the generative model and dashed lines represent additional computation for inference. }
\end{figure}

\subsection{Generative model}

Given the observations for a multi-object dynamical system, our model further
assumes its interaction structure is known as prior knowledge. The interaction
structure is provided as a directed graph, in which each object corresponds to
a vertex, and a directed edge indicates that the state of its head is
likely to be affected by its tail. In situations where such graph structure
is not available, a complete graph can be specified. However, to model
dynamical systems comprising a large number of objects, it is often beneficial
to explicitly specify sparse graph structures, because they impose stronger
relational inductive bias and help save the computational cost.

A relational state-space model assumes a set of correlated dynamical subsystems evolve
jointly under the coordination of graph neural networks. Formally, given a
graph $\CG = (\CV, \CE, \MV, \TE)$, in which an edge $(i, j) \in
\CE$ indicates that the state of vertex $j$ may be affected by vertex $i$. Let
$\Vu_{t}^{(i)} \in \BR^{d_u}$ and $\Vx_{t}^{(i)} \in \BR^{d_x}$ be the input and observation for vertex $i$ at time step $t$, respectively. For $T$ steps, we introduce a set of unobserved random variables $\{\Vz_{1:T}^{(i)}\}_{i=1}^N$, in which
$\Vz_{t}^{(i)} \in \BR^{d_z}$ represents the latent state of vertex $i$ at time
step $t$. Furthermore, we introduce a global latent variable $\Vz_t^g \in \BR^{d_g}$ for each time step to represent the global state shared by all vertices. Conditioning on the graph and exogenous inputs, an R-SSM factorizes the joint density of observations and latent states as follows:
\begin{multline}  \label{eq:r-ssm-joint-density}
p_{\theta} \left(
	\big\{ \Vx_{1:T}^{(i)} \big\}_{i=1}^N,
	\big\{ \Vz_{1:T}^{(i)} \big\}_{i=1}^N, \Vz_{1:T}^g
	\middle|
	\CG, \big\{ \Vu_{1:T}^{(i)} \big\}_{i=1}^N
\right)
= \\
\prod_{t=1}^{T}{
	f_{\theta}\left(
		\big\{ \Vz_t^{(i)} \big\}_{i=1}^N, \Vz_t^g
		\middle|
		\big\{ \Vz_{<t}^{(i)} \big\}_{i=1}^N, \Vz_{<t}^g,
		\CG, \big\{ \Vu_{\leq t}^{(i)} \big\}_{i=1}^N
	\right)
} \\
\prod_{t=1}^{T}{\prod_{i=1}^{N}{
		g_{\theta} \left(
		\Vx_t^{(i)}
		\middle|
		\Vz_t^{(i)}, \Vz_{\leq t}^g, \big\{ \Vz_{<t}^{(i)} \big\}_{i=1}^N,
		\CG, \big\{ \Vu_{\leq t}^{(i)} \big\}_{i=1}^N
		\right)
}}
\end{multline}
For notational simplicity, we switch to the matrix notation $\MZ_t =
\big[\Vz_t^{(1)}, \Vz_t^{(2)}, \ldots, \Vz_t^{(N)} \big]^{\top}$ from now on.
The joint transition density $f_{\theta}$ is further factorized as a product of global transition density $f_{\theta}^g$ and local transition density $f_{\theta}^{\star}$, i.e.,
\(
f_{\theta}\left(\MZ_t, \Vz_t^g \middle| \ldots \right) =
	f_{\theta}^g\left(\Vz_t^g \middle| \ldots \right)
	f_{\theta}^{\star}\left(\MZ_t \middle| \Vz_t^g, \ldots \right)
\)
. To instantiate these conditional distributions, a
GNN accompanied by RNN cells is adopted to recurrently compress the past
dependencies at each time step into fixed-size context
vectors. Specifically, the observations are assumed to be generated from
following process:
\begin{align*}
\tilde{\Vh}_t^{(i)} =
\text{RNN}_{\theta}^v\big( \Vh_{t-1}^{(i)}, \big[ \Vz_{t-1}^{(i)}, \Vu_t^{(i)} \big] \big)
, \quad &
\tilde{\Vh}_t^g = \text{READOUT}_{\theta}\big( \CG, \tilde{\MH}_t \big)
\\
\Vh_t^g =
\text{RNN}_{\theta}^g\big( \Vh_{t-1}^g, \big[ \Vz_{t-1}^g, \tilde{\Vh}_t^g \big] \big)
, \quad &
\Vz_t^g \sim f_{\theta}^g\big( \cdot \big| \Vh_t^g \big)
\\
\MH_t = \text{GNN}_{\theta}\big( \CG, \Vz_t^g, \tilde{\MH}_t \big)
, \quad &
\MZ_t \sim
f_{\theta}^{\star} \big( \cdot \big| \MH_t \big)
\\
& \Vx_t^{(i)} \sim
g_{\theta}\big( \cdot \big| \Vz_t^{(i)}, \Vh_t^{(i)}, \Vz_t^g, \Vh_t^g \big)
\end{align*}
for $i = 1, \ldots, N$ and $t = 1, \ldots, T$, where $\Vh_0^{(i)} =
\Vh_0^{\ast}$ and $\Vz_0^{(i)} = \Vz_0^{\ast}$. Here $\Vh_0^g$, $\Vz_0^g$,
$\Vh_0^{\ast}$ and $\Vz_0^{\ast}$ are learnable initial states. The READOUT
function aggregates the context vectors of all vertices into a global context
vector in a permutation-invariant manner. The global transition density
$f_{\theta}^g$ is specified to be a diagonal Gaussian distribution whose mean
and variance are parameterized by the output of a multilayer perceptron (MLP),
and the local transition density $f_{\theta}^{\star}$ will be discussed later.
The local observation distribution $g_{\theta}$ can be freely selected in line with
the data, and in our experiments it is either a Gaussian distribution or a mixture
of logistic distributions parameterized by MLPs.

The graphical structure of two consecutive steps of the generating process is
illustrated in Figure \ref{fig:gen-model}. An intuitive way to think about
this generative model is to note that the $N + 1$ latent state processes
interact through the GNN, which enables the new state of a vertex to depend on
not only its own state trajectory but also the state trajectories of other
vertices and the entire graph.

\subsection{Learning and inference}

As illustrated in Figure \ref{fig:gen-model-collapsed}, writing $\CZ_t = (\Vz_t^g, \MZ_t)$ and suppressing the dependencies on the graph $\CG$ and exogenous inputs $\MU_{1:T}$, an R-SSM can be interpreted as an ordinary SSM in which the entire graph evolves as a whole, i.e., the joint density of latent states and observations factors as:
$
p(\MX_{1:T}, \CZ_{1:T}) =
\prod_{t=1}^{T}{p_{\theta}(\CZ_t|\CZ_{<t}) p_{\theta}(\MX_t|\CZ_t)}
$.
Given observations $\MX_{1:T}$, we are interested in learning unknown
parameters $\theta$ and inferring unobserved states $\CZ_{1:T}$. For the
learning task we wish to maximize the marginal likelihood
$
p_{\theta}(\MX_{1:T}) =
\int p_{\theta}(\MX_{1:T}, \CZ_{1:T}) \: \mathrm{d}\CZ_{1:T}
$, 
but in our case the integral is intractable. We adopt a
recently developed variational inference (VI) approach called variational sequential Monte Carlo (VSMC) \cite{vsmc-1,vsmc-2,vsmc-3}, which
maximizes a variational lower bound on the log marginal likelihood instead and
learns the proposal distributions for the inference task simultaneously.

Given a sequence of proposal distributions
$\{q_{\phi}\left(\CZ_t \middle| \CZ_{<t}, \MX_{\leq t} \right)\}_{t=1}^T$
parameterized by $\phi$, running the sequential Monte Carlo
(SMC) algorithm with $K$ particles yields an unbiased
marginal likelihood estimator
$
\hat{p}_{\theta,\phi,K}(\MX_{1:T}) =
\prod_{t=1}^{T}{\big[ \nicefrac{1}{K} \sum_{k=1}^{K}{w_t^k} \big]}
$,
where $w_t^k$ is the unnormalized importance weight of particle $k$ at time
 $t$. The variational lower bound is obtained by applying the Jensen's
inequality:
\begin{equation}
\CL_K^{\text{SMC}}(\theta, \phi)
\coloneqq \BE \left[ \log \hat{p}_{\theta,\phi,K}(\MX_{1:T}) \right ]
\leq \log \BE \left[ \hat{p}_{\theta,\phi,K}(\MX_{1:T}) \right ]
= \log p_{\theta}(\MX_{1:T})
\end{equation}
Assuming the proposal distributions are reparameterizable \cite{vae}, we use the
biased gradient estimator
$
\nabla \CL_K^{\text{SMC}}(\theta, \phi) \approx
\BE[\nabla \log \hat{p}_{\theta, \phi, K}(\MX_{1:T})]
$
to maximize $\CL_K^{\text{SMC}}$.

\paragraph{Proposal design.} We make the proposal for $\CZ_t$
depend on the information up to time $t$ and share some parameters with the generative model. We also choose to factorize
\(
q_{\phi}\left(\CZ_t \middle| \ldots \right) =
r_{\phi}^g \left( \Vz^g_t \middle| \ldots \right)
r_{\phi}^{\star} \left( \MZ_t \middle| \Vz^g_t, \ldots \right)
\)
. The proposal distributions for all time steps are structured as follows:
\begin{align*}
\tilde{\Vb}_t^{(i)} =
\text{RNN}_{\phi}\big( \Vb_{t-1}^{(i)}, \big[ \Vx_t^{(i)}, \Vu_t^{(i)} \big] \big)
, \quad &
\tilde{\Vb}_t^g = \text{READOUT}_{\phi, 1}\big(\CG, \tilde{\MB}_t \big)
\\
\MB_t = \text{GNN}_{\phi}\big( \CG, \tilde{\Vb}_t^g, \tilde{\MB}_t \big)
, \quad &
\Vb_t^g = \text{READOUT}_{\phi, 2}\big(\CG, \MB_t \big)
\\
\Vz_t^g \sim r_{\phi}^g\big( \cdot \big| \Vh_t^g, \Vb_t^g \big)
, \quad &
\MZ_t \sim
r_{\phi}^{\star} \big( \cdot \big| \MH_t, \MB_t \big)
\end{align*}
for $i = 1, \ldots, N$ and $t = 1, \ldots, T$, where $\Vh_t^g$ and $\MH_{1:T}$
are computed using the relevant parts of the generative model. $r_{\phi}^g$ is
specified to be a diagonal Gaussian parameterized by an MLP, and
$r_{\phi}^{\star}$ will be discussed soon. Here $\MB_t$ can be interpreted as a
\emph{belief state} \cite{td-vae}, which summarizes past observations
$\MX_{\leq t}$ (and inputs $\MU_{\leq t}$) deterministically. The graphical
structure of this proposal design is shown in Figure \ref{fig:proposal}, and the
detailed VSMC implementation using this proposal is given in Appendix
\ref{subsec:app-train}.

\subsection{Graph Normalizing Flow}

The local transition density $f_{\theta}^{\star} \left( \MZ_t \big| \ldots
\right)$ and the local proposal density $r_{\phi}^{\star} \left( \MZ_t \middle|
\ldots \right)$ may be parameterized in several ways. One simple and efficient
starting point is (block-)diagonal Gaussian distribution: $f_{\theta}^{\star}
\left( \MZ_t \big| \ldots \right) = \prod_{i=1}^{N}\CN(\Vz_t^{(i)} | \ldots )$,
which assumes that the object states are conditionally independent, i.e., the
joint state distribution is completely factorized over objects. We believe that
such an independence assumption is an oversimplification for situations where
the joint state evolution is multimodal and highly correlated. One possible way
to introduce inter-object dependencies is modeling joint state distributions as
Markov random fields (MRFs) \cite{nested-smc}, but this will significantly
complicate the learning.  

Here we introduce the Graph Normalizing Flow (GNF)
\footnote{GNF has been independently developed by \citet{gnf} for different purpose.}, which adapts Glow \cite{glow} to graph settings and enables us to build expressive joint distributions for correlated random variables indexed by graph nodes. As described earlier, the key ingredient for a flow is a series invertible mappings that are iteratively applied to the samples of a base distribution. Now we are interested in the case where the samples are vertex states $\MZ_t$, and thus the invertible mappings should be further constrained to be \emph{equivariant} under vertex relabeling. This rules out popular autoregressive flows, e.g., IAF \cite{iaf} and MAF \cite{maf}.

Our GNF is built upon the \emph{coupling} layer introduced in \citet{real-nvp}, which provides a flexible framework to construct efficient invertible mappings. A GNF coupling layer splits the input $\MZ \in \BR^{N \times D}$ into two parts, $\MZ_a \in \BR^{N \times d}$ and $\MZ_b \in \BR^{N \times (D - d)}$. The output $\MZ' \in \BR^{N \times D}$ is formed as:
\[
\MZ'_a = \MZ_a
\; , \quad
\MZ'_b = \MZ_b \odot \exp\big(s(\MZ_a)\big) + t(\MZ_a)
\; , \quad
\MZ' = [\MZ'_a, \MZ'_b] \; ,
\]
where $\odot$ denotes the element-wise product, and the functions $s(\cdot)$
and $t(\cdot)$ are specified to be GNNs to enforce the equivariance property.
A GNF combines a coupling layer with a trainable element-wise affine layer and an invertible $1 \times 1$ convolution layer \cite{emerging}, organizing them as:
\(
\text{Input}
\rightarrow \text{Affine}
\rightarrow \text{Coupling}
\rightarrow \text{Conv}_{1\times1}
\rightarrow \text{Output}
\).
A visual illustration of this architecture is provided in Appendix \ref{subsec:app-gnf}.

In order to obtain more expressive prior and variational posterior
approximation, the local transition density and local proposal density can be
constructed by stacking multiple GNFs on top of diagonal Gaussian
distributions parameterized by MLPs. With the message passing inside the
coupling layers, GNFs can transform independent noise into correlated noise
and thus increase model expressivity. The $1 \times 1$ convolution layers free
us from manually permuting the dimensions, and the element-wise affine layers
enable us to tune their initial weights to stablize training.

\subsection{Auxiliary contrastive prediction tasks}

In our initial experiments, we found that learning R-SSM suffered from the
\emph{posterior collpase} phenomenon, which is a well known problem in the
training of variational autoencoders (VAEs). It means that the variational posterior
approximation $q_{\phi}(\CZ_t|\CZ_{<t}, \MX_{\leq t})$ degenerate into the
prior $f_{\theta}(\CZ_t|\CZ_{<t})$ in the early stage of optimization, making
the training dynamics get stuck in undesirable local optima. Besides, we also
encountered a more subtle problem inherent in likelihood-based training of
deep sequential models. That is, for relatively smooth observations, the
learned model tended to only capture short-term local correlations but not
the interaction effects and long-term transition dynamics.

Motivated by recent advances in unsupervised representation learning based on
mutual information maximization, in particular the Contrastive Predictive
Coding (CPC) approach \cite{cpc}, we alleviate these problems by forcing the latent
states to perform two auxiliary contrastive prediction tasks.
At each time step $t$, the future observations of each vertex $i$ are
summarized into a vector using a backward RNN: $\Vc_t^{(i)} =
\text{RNN}_{\psi}(\Vx_{>t}^{(i)} )$. Then we define two auxiliary CPC
objectives:
\[
\CL_1^{\text{aux}} = \BE \left[ \sum_{t=1}^{T-1} \sum_{i=1}^N
  \log \frac{
    \lambda_{\psi, 1}( \hat{\Vz}_t^{(i)}, \Vc_t^{(i)} )
  }{
    \sum_{ \Vc \in \Omega_{t,i} }{
       \lambda_{\psi, 1}( \hat{\Vz}_t^{(i)}, \Vc ) 
    }
  }
\right]
, \;
\CL_2^{\text{aux}} = \BE \left[ \sum_{t=1}^{T-1} \sum_{i=1}^N
  \log \frac{
    \lambda_{\psi, 2}( \hat{\Vh}_t^{(i)}, \Vc_t^{(i)} )
  }{
    \sum_{ \Vc \in \Omega_{t,i} }{
       \lambda_{\psi, 2}( \hat{\Vh}_t^{(i)}, \Vc ) 
    }
  }
\right]
\]
where
$\hat{\Vz}_t^{(i)} = [\Vz_t^g, \Vz_t^{(i)}]$, 
$\hat{\Vh}_t^{(i)} = \text{MLP}_{\psi}( \sum_{ j \neq i \, \wedge \, j \in \CN_i^{-} }{ \Vh_t^{(j)} } )$, and $\Omega_{t,i}$ is a set that contains $\Vc_t^{(i)}$ and some negative samples. The expectation is over negative samples and the latent states sampled from the filtering distributions. The positive score functions $\lambda_{\psi, 1}$ and $\lambda_{\psi, 2}$ are specified to be simple log-bilinear models.

Intuitively, $\CL_1^{\text{aux}}$ encourages the latent states to encode
useful information that helps distinguish the future summaries from negative
samples. $\CL_2^{\text{aux}}$ encourages the deterministic states to reflect
the interaction effects, as it contrastingly predicts the future summary of
vertex $i$ based on the states of $i$'s neighbors only. The negative samples
are selected from the future summaries of other vertices within the minibatch.
The final objective to maximize is
$\CL = \CL_K^{\text{SMC}} + \beta_1 \CL_1^{\text{aux}} + \beta_2 \CL_2^{\text{aux}}$, in which $\beta_1 \geq 0$ and $\beta_2 \geq 0$ are tunable hyperparameters. The procedure to estimate this objective is described in Appendix \ref{subsec:app-train}.

%% file: img/generative-model.tex
\begin{tikzpicture}
[
visible/.style={
	shape=circle,draw=blue!50,fill=blue!20,thick,
	node font=\Huge,inner sep=0pt,minimum size=2.25cm
},
latent/.style={
	shape=circle,draw,thick,
	node font=\Huge,inner sep=0pt,minimum size=2.25cm
},
hidden/.style={
	shape=diamond,draw,thick,
	node font=\Huge,inner sep=0pt,minimum size=3.0cm
},
arrow/.style={-{Latex[length=4mm]},thick}
]

\node[hidden] (H_t)                       {$\mathbf{H}_t$};
\node[hidden] (H_tmp)    [right=of H_t]   {$\tilde{\mathbf{H}}_{t+1}$};
\node[hidden] (H_next)   [right=of H_tmp] {$\mathbf{H}_{t+1}$};

\node[latent] (z_g_t)    [above=of H_t]      {$\mathbf{z}^g_t$};
\node[latent] (z_g_next) [above=of H_next]   {$\mathbf{z}^g_{t+1}$};

\node[hidden] (h_g_t)    [above=of z_g_t]       {$\mathbf{h}^g_t$};
\node[hidden] (h_g_next) [above=of z_g_next]    {$\mathbf{h}^g_{t+1}$};

\draw[arrow]  (h_g_t)    to (h_g_next) ;

\draw[arrow]  (h_g_t)    to (z_g_t)    ;
\draw[arrow]  (h_g_next) to (z_g_next) ;

\draw[arrow]  (z_g_t)    to (h_g_next) ;

\draw[arrow]  (H_t)      to (H_tmp)    ;
\draw[arrow]  (H_tmp)    to (H_next)   ;
\draw[arrow]  (H_tmp.north) to (h_g_next) ;

\node[latent] (Z_t)    [below=of H_t]    {$\mathbf{Z}_t$};
\node[latent] (Z_next) [below=of H_next] {$\mathbf{Z}_{t+1}$};

\draw[arrow]  (H_t)    to (Z_t)    ;
\draw[arrow]  (H_next) to (Z_next) ;

\draw[arrow]  (Z_t)    to (H_tmp) ;

\node[visible] (X_t)    [below=of Z_t]    {$\mathbf{X}_t$};
\node[visible] (X_next) [below=of Z_next] {$\mathbf{X}_{t+1}$};

\draw[arrow]  (Z_t)    to (X_t)    ;
\draw[arrow]  (Z_next) to (X_next) ;

\draw[arrow]  (z_g_t)    to (H_t)    ;
\draw[arrow]  (z_g_next) to (H_next) ;

\draw[arrow]  (h_g_t.south west)    to [bend right=45] (X_t.north west)   ;
\draw[arrow]  (h_g_next.south east) to [bend left=45] (X_next.north east) ;

\draw[arrow]  (z_g_t.south west)    to [bend right=41] (X_t.north west)   ;
\draw[arrow]  (z_g_next.south east) to [bend left=41] (X_next.north east) ;

\draw[arrow]  (H_t.south west)    to [bend right=42] (X_t.north west)   ;
\draw[arrow]  (H_next.south east) to [bend left=42] (X_next.north east) ;

\end{tikzpicture}

%% file: img/generative-model-collapsed.tex
\begin{tikzpicture}
[
visible/.style={
	shape=circle,draw=blue!50,fill=blue!20,thick,
	node font=\Huge,inner sep=0pt,minimum size=2.0cm
},
latent/.style={
	shape=circle,draw,thick,
	node font=\Huge,inner sep=0pt,minimum size=2.0cm
},
hidden/.style={
	shape=diamond,draw,thick,
	node font=\Huge,inner sep=0pt,minimum size=2.0cm
},
arrow/.style={-{Latex[length=4mm]},thick}
]

\node[latent] (Z_1)                      {$\mathcal{Z}_1$};
\node[latent] (Z_2)    [right=of Z_1]    {$\mathcal{Z}_2$};
\node[latent] (Z_3)    [right=of Z_2]    {$\mathcal{Z}_3$};
\node[latent] (Z_dots) [right=of Z_3]    {$\cdots$};
\node[latent] (Z_T)    [right=of Z_dots] {$\mathcal{Z}_T$};

\draw[arrow]  (Z_1)    to (Z_2)    ;
\draw[arrow]  (Z_2)    to (Z_3)    ;
\draw[arrow]  (Z_3)    to (Z_dots) ;
\draw[arrow]  (Z_dots) to (Z_T)    ;

\node[visible] (X_1)    [below=2cm of Z_1]  {$\mathbf{X}_1$};
\node[visible] (X_2)    [right=of X_1]      {$\mathbf{X}_2$};
\node[visible] (X_3)    [right=of X_2]      {$\mathbf{X}_3$};
\node[visible] (X_dots) [right=of X_3]      {$\cdots$};
\node[visible] (X_T)    [right=of X_dots]   {$\mathbf{X}_T$};

\draw[arrow]  (Z_1)    to (X_1)    ;
\draw[arrow]  (Z_2)    to (X_2)    ;
\draw[arrow]  (Z_3)    to (X_3)    ;
\draw[arrow]  (Z_dots) to (X_dots) ;
\draw[arrow]  (Z_T)    to (X_T)    ;

\draw[arrow]  (Z_1.north east)    to [bend left=45] (Z_3.north west)    ;
\draw[arrow]  (Z_1.north east)    to [bend left=45] (Z_dots.north west) ;
\draw[arrow]  (Z_1.north east)    to [bend left=45] (Z_T.north west)    ;
\draw[arrow]  (Z_2.north east)    to [bend left=45] (Z_dots.north west) ;
\draw[arrow]  (Z_2.north east)    to [bend left=45] (Z_T.north west)    ;
\draw[arrow]  (Z_3.north east)    to [bend left=45] (Z_T.north west)    ;

\end{tikzpicture}

%% file: img/proposal.tex
\begin{tikzpicture}
[
visible/.style={
	shape=circle,draw=blue!50,fill=blue!20,thick,
	node font=\Huge,inner sep=0pt,minimum size=2.25cm
},
latent/.style={
	shape=circle,draw,thick,
	node font=\Huge,inner sep=0pt,minimum size=2.25cm
},
hidden/.style={
	shape=diamond,draw,thick,
	node font=\Huge,inner sep=0pt,minimum size=3.0cm
},
arrow/.style={-{Latex[length=4mm]},thick},
new arrow/.style={-{Latex[length=5mm]},thick,densely dashed,red}
]

\node[hidden] (H_t)                       {$\mathbf{H}_t$};
\node[hidden] (H_tmp)    [right=of H_t]   {$\tilde{\mathbf{H}}_{t+1}$};
\node[hidden] (H_next)   [right=of H_tmp] {$\mathbf{H}_{t+1}$};

\node[latent] (z_g_t)    [above=of H_t]      {$\mathbf{z}^g_t$};
\node[latent] (z_g_next) [above=of H_next]   {$\mathbf{z}^g_{t+1}$};

\node[hidden] (h_g_t)    [above=of z_g_t]       {$\mathbf{h}^g_t$};
\node[hidden] (h_g_next) [above=of z_g_next]    {$\mathbf{h}^g_{t+1}$};

\draw[arrow]  (h_g_t)    to (h_g_next) ;

\draw[arrow]  (h_g_t)    to (z_g_t)    ;
\draw[arrow]  (h_g_next) to (z_g_next) ;

\draw[arrow]  (z_g_t)    to (h_g_next) ;

\draw[arrow]  (H_t)      to (H_tmp)    ;
\draw[arrow]  (H_tmp)    to (H_next)   ;
\draw[arrow]  (H_tmp.north) to (h_g_next) ;

\node[latent] (Z_t)    [below=of H_t]    {$\mathbf{Z}_t$};
\node[latent] (Z_next) [below=of H_next] {$\mathbf{Z}_{t+1}$};

\draw[arrow]  (H_t)    to (Z_t)    ;
\draw[arrow]  (H_next) to (Z_next) ;

\draw[arrow]  (Z_t)    to (H_tmp) ;

\node[hidden]  (B_t)    [below=of Z_t]    {$\mathbf{B}_t$};
\node[hidden]  (B_next) [below=of Z_next] {$\mathbf{B}_{t+1}$};

\draw[new arrow]  (B_t)    to (Z_t)    ;
\draw[new arrow]  (B_next) to (Z_next) ;

\draw[new arrow]  (B_t) to (B_next) ;

\draw[new arrow] (B_t.north west)    to [bend left=45] (z_g_t.south west);
\draw[new arrow] (B_next.north east) to [bend right=45] (z_g_next.south east);

\node[visible] (X_t)    [left=of B_t]     {$\mathbf{X}_t$};
\node[visible] (X_next) [right=of B_next] {$\mathbf{X}_{t+1}$};

\draw[new arrow]  (X_t)    to (B_t)    ;
\draw[new arrow]  (X_next) to (B_next) ;

\draw[arrow]  (z_g_t)    to (H_t)    ;
\draw[arrow]  (z_g_next) to (H_next) ;

\end{tikzpicture}

%% file: tex/related.tex
\section{Related Work}

\paragraph{GNN-based dynamics modeling.} GNNs
\cite{gnn-09,gcn-15,ggnn,gcn-16,mpnn,graph-sage,gat,gnn-power,eq-gn} provide a
promising framework to learn on graph-structured data and impose relational
inductive bias in learning models. We refer the reader to \citet{gn} for a
recent review. GNNs (or neural message passing modules) are the core
components of recently developed neural physics simulators
\cite{in,vin,obj-1,obj-2,gn-icml,lifeifei-1,fluid} and spatiotemporal or
multi-agent dynamics models
\cite{social-lstm,vain,dcrnn,gaan,r-forward,chen-aaai2020}. In these works, GNNs
usually act autoregressively or be integrated into the sequence-to-sequence
(seq2seq) framework \cite{seq2seq}. Besides, recently they have been combined
with generative adversarial networks \cite{gan} and normalizing flows for multi-agent forecasting \cite{social-gan,social-bigat,precog}. R-SSM differs from all
these works by introducing structured latent variables to represent the
uncertainty on state transition and estimation.

\paragraph{GNNs in sequential LVMs.}
A few recent works have combined GNNs with a sequential latent variable model,
including R-NEM \cite{r-nem}, NRI \cite{nri}, SQAIR \cite{seq-air}, VGRNN
\cite{vgrnn}, MFP \cite{mfp}, and Graph VRNN \cite{graph-vrnn,graph-vrnn-cvpr}. The latent
variables in R-NEM and NRI are discrete and represent membership relations and
types of edges, respectively. In contrast, the latent variables in our model
are continuous and represent the states of objects. SQAIR is also a deep SSM
for multi-object dynamics, but the GNN is only used in its inference network.
VGRNN is focused on modeling the topological evolution of dynamical graphs.
MFP employs a conditional VAE architecture, in which the per-agent discrete latent
variables are shared by all time steps.
The work most relevant to ours is Graph VRNN, in which the hidden states of
per-agent VRNNs interact through GNNs. Our work mainly differs from it by
introducing a global latent state process to make the model hierarchical and
exploring the use of normalizing flows as well as the auxiliary contrastive
objectives. More subtle differences are discussed in Section \ref{subsec:ball}.

\paragraph{Deep LVMs for sequential data.} There has been growing
interest in developing latent variable models for sequential data with neural
networks as their building blocks, among which the works most relevant to ours
are stochastic RNNs and deep SSMs. Many works have proposed incorporating
stochastic latent variables into vanilla RNNs to equip them with the ability
to express more complex data distributions \cite{storn,vrnn,srnn,zf,zf-rl} or,
from another perspective, developing deep SSMs by parameterizing flexible
transition and emission distributions using neural networks
\cite{dmm,kvae,s-ssm,ss-lstm,rssm}. Approximate inference and parameter
estimation methods for nonlinear SSMs have been extensively studied in the
literature \cite{smc-tutorial,pmcmc,particle,nasmc,vbf,avf,td-vae,vb-smc}. We
choose VSMC \cite{vsmc-1,vsmc-2,vsmc-3} as it combines the powers of VI and
SMC. The posterior collapse problem is commonly addressed by KL annealing,
which does not work with VSMC. The idea of using auxiliary costs to train deep
SSMs has been explored in Z-forcing \cite{zf,zf-rl}, which predicts the future
summaries directly rather than contrastingly. As a result, the backward RNN in
Z-forcing may degenerate easily.

%% file: tex/exp.tex
\section{Experiments}

We implement R-SSM using the TensorFlow Probability library \cite{tfd}. The
experiments are organized as follows: In Section \ref{subsec:toy}, we sample a
toy dataset from a simple stochastic multi-object model and validate that R-SSM
can fit it well while AR models and non-relational models may struggle. In
Section \ref{subsec:ball}, R-SSM is compared with state-of-the-art sequential
LVMs for multi-agent modeling on a basketball gameplay dataset, and the
effectiveness of GNF is tested through ablation studies. Finally,
in Section \ref{subsec:traffic}, the prediction performance of R-SSM is
compared with strong GNN-based seq2seq baselines on a road traffic dataset. Due
to the space constraint, the detailed model architecture and hyperparameter
settings for each dataset are given in the Appendix. Below, all values reported
with error bars are averaged over 3 or 5 runs.

\subsection{Synthetic toy dataset} \label{subsec:toy}

First we construct a simple toy dataset to illustrate the capability of R-SSM.
Each example in this dataset is generated by the following procedure:
\begin{gather}
\CG \sim \text{SBM}(N, K, p_0, p_1)
, \quad
\Vv_i \sim \text{Normal}(\V{0}, \M{I})
, \quad
z_0^{(i)} \sim \text{Normal}(0, 1)
\nonumber
\\
\label{eq:toy}
\tilde{z}_t^{(i)} = \bm{\eta}^{\top} \Vv_i  +
	\alpha_1 {\textstyle \sum}_{j \in \CN_i}^{}{z_{t-1}^{(j)}} / |\CN_i| +
	\alpha_2 z_{t-1}^{(i)}
\\
z_t^{(i)} \sim \text{Normal}\big( \cos( \tilde{z}_t^{(i)} ), \sigma_z^2 \big)
, \quad
x_t^{(i)} \sim \text{Normal}\big( \tanh( \varepsilon z_t^{(i)} ), \sigma_x^2 \big)
\nonumber
\end{gather}
for $i = 1, \ldots, N$ and $t = 1, \ldots, T$. Here $\text{SBM}$ is short for
the symmetric stochastic block model, in which each vertex $i$ belongs to
exact one of the $K$ communities, and two vertices $i$ and $j$ are connected
with probability $p_0$ if they are in the same community, $p_1$ otherwise. A
vertex-specific covariate vector $\Vv_i \in \BR^{d_v}$ is attached to each
vertex $i$, and by Equation \eqref{eq:toy}, the state of each vertex $i$ can be
affected by its neighbors $\CN_i$. Choosing the parameters $d_v = 4$, $N =
36$, $K = 3$, $p_0 = 1/3$, $p_1 = 1/18$, $T = 80$, $\alpha_1 = 5.0$, $\alpha_2
= -1.5$, $\bm{\eta} = [ -1.5, 0.4, 2.0, -0.9 ]^{\top}$, $\sigma_x = \sigma_z =
0.05$, and $\varepsilon = 2.5$, we generate 10K examples for training, validation, and
test, respectively. A typical example is visualized in the Appendix.

Despite the simple generating process, the resulting dataset is highly challenging for common models to fit. To show this, we compare R-SSM with several baselines, including
\begin{enumerate*}[label={(\alph*)}]
  \item VAR: Fitting a first-order vector autoregression model for each example; 
  \item VRNN: A variational RNN \cite{vrnn} shared by all examples;
  \item GNN-AR: A variant of the recurrent decoder of NRI \cite{nri}, which is exactly a GNN-based AR model when given the ground-truth graph.
\end{enumerate*}
VAR and VRNN are given access to the observations $\{ x_{1:T}^{(i)} \}_{i=1}^N$ only, while GNN-AR and R-SSM are additionally given access to the graph structure $(\CV, \CE)$ (but not the vertex covariates). GNF is not used in R-SSM because the true joint transition distribution is factorized over vertices.

\begin{wraptable}{r}{0.525\textwidth}
\centering
\caption{Test log-likelihood and prediction performance comparisons on the synthetic toy dataset.}\label{tbl:toy}
\input{data/toy/metrics}
\end{wraptable}
For each model, we calculate three metrics:
\begin{enumerate*}[label={(\arabic*)}]
  \item LL: Average log-likelihood (or its lower bound) of test examples; 
  \item MSE: Average mean squared one-step prediction error given the first 75 time steps of each test example;
  \item CP: Average coverage probability of a 90\% one-step prediction interval.
\end{enumerate*}
For non-analytic models, point predictions and prediction intervals are computed using 1000 Monte Carlo samples. The results are reported in Table \ref{tbl:toy}.

The generating process involves latent factors and nonlinearities, so VAR
performs poorly as expected. VRNN largely underfits the data and struggles to
generalize, which may be caused by the different topologies under the
examples. In contrast, GNN-AR and R-SSM generalize well as expected, while
R-SSM achieves much higher test log-likelihood and produces good one-step
probabilistic predictions. This toy case illustrates the generalization
ability of GNNs and suggests the importance of latent variables for capturing
the uncertainty in stochastic multi-object systems. We also observed that
without $\CL_1^{\text{aux}}$ the training dynamics easily get stuck in
posterior collapse at the very early stage, and adding $\CL_2^{\text{aux}}$
help improve the test likelihood.

\begin{table}
  \centering
  \caption{
  Test log-likelihood and rollout quality comparisons on the basketball
	gameplay dataset (offensive players only).
  }
  \input{data/basketball/metrics}
  \label{tbl:player}
\end{table}

\subsection{Basketball gameplay}\label{subsec:ball}

In basketball gameplay, the trajectories of players and the ball are highly
correlated and demonstrate rich, dynamic interations. Here we compare R-SSM
with a state-of-the-art hierarchical sequential LVM for multi-agent
trajectories \cite{weak-sup}, in which the per-agent VRNNs are coordinated by
a global "macro intent" model. We note it as MI-VRNN. The dataset\footnote{Data Source: STATS, copyright 2019.} includes
107,146 training examples and 13,845 test examples, each of which contains the
2D trajectories of ten players and the ball recorded at 6Hz for 50 time steps.
Following their settings, we use the trajectories of offensive team only and
preprocess the data in exactly the same way to make the results directly
comparable. The complete graph of players is used as the input to R-SSM.

Several ablation studies are performed to verify the utility of the proposed
ideas. In Table \ref{tbl:ball}, we report test likelihood bounds and the
rollout quality evaluated with three heuristic statistics: average speed
(feet/step), average distance traveled (feet), and the percentage of
out-of-bound (OOB) time steps. The VRNN baseline developed by \citet{weak-sup}
is also included for comparison. Note that the VSMC bound
$\CL_{1000}^{\text{SMC}}$ is a tighter log-likelihood approximation than the
ELBO (which is equivalent to $\CL_{1}^{\text{SMC}}$). The rollout statistics
of R-SSMs are calculated from 150K 50-step rollouts with 10 burn-in steps. Several
selected rollouts are visualized in the Appendix.

As illustrated in Table \ref{tbl:player}, all R-SSMs outperform the baselines in
terms of average test log-likelihood. Again, we observed that adding
$\CL_1^{\text{aux}}$ is necessary for training R-SSM successfully on this
dataset. Training with the proposed auxiliary loss $\CL_2^{\text{aux}}$ and
adding GNFs do improve the results. R-SSM with 8 GNFs (4 in prior, 4 in
proposal) achieves higher likelihood than R-SSM with 4 GNFs, indicating that
increasing the expressivity of joint state distributions helps fit the data
better. As for the rollout quality, the OOB rate of the rollouts sampled from
our model matches the ground-truth significantly better, while the other two
statistics are comparable to the MI-VRNN baseline. 

In Table \ref{tbl:ball}, we also provide preliminary results for the setting
that additionally includes the trajectory of the ball. This enables us to
compare with the results reported by \citet{graph-vrnn-cvpr} for Graph VRNN
(GVRNN).
The complete graph of ball and players served as input to R-SSM is annotated
with two node types (player or ball) and three edge types (player-to-ball,
ball-to-player or player-to-player). R-SSM achieves competitive test
likelihood, and adding GNFs helps improve the performance. We point out that
several noticeable design choices of GVRNN may help it outperform R-SSM: 
\begin{enumerate*}[label={(\roman*)},font={\bfseries}]
\item GVRNN uses a GNN-based observation model, while R-SSM uses a simple
	factorized observation model.
\item GVRNN encodes $\MX_{1:t-1}$ into $\MH_t$ and thus enables the prior of
	$\MZ_t$ to depend on past observations, which is not the case in R-SSM.
\item GVRNN uses several implementation tricks, e.g., predicting the changes in
	observations only ($\MX_t = \MX_{t-1} + {\Delta \MX}_t$) and passing raw
		observations as additional input to GNNs.
\end{enumerate*}
We would like to investigate the effect of these interesting differences in future work.

\begin{table}
\parbox{.45\linewidth}{
  \centering
  \caption{
	  Test log-likelihood comparison on the basketball gameplay dataset (offensive players
	  plus the ball).
  }
  \input{data/basketball/ball-metrics}
  \label{tbl:ball}
}
\hfill
\parbox{.5\linewidth}{
  \centering
  \caption{Forecast MAE comparison on the METR-LA dataset. $h$ is the number of steps predicted into the future. The $\MX_{t-h}$ baseline outputs $\MX_{t-h}$ to predict $\MX_t$.}\label{tbl:traffic-mae}
  \input{data/traffic/table-mae}
  \label{tbl:traffic-mae}
}
\end{table}

\subsection{Road traffic}\label{subsec:traffic}

Traffic speed forecasting on road networks is an important but challenging
task, as the traffic dynamics exhibit complex spatiotemporal interactions. In
this subsection, we demonstrate that R-SSM is comparable to the
state-of-the-art GNN-based seq2seq baselines on a real-world
traffic dataset. The METR-LA dataset \cite{dcrnn} contains 4 months of
1D traffic speed measurements that were recorded via 207 sensors and aggregated
into 5 minutes windows. For this dataset, all conditional inputs $\CG = (\CV,
\CE, \MV, \TE)$ and $\MU_{1:T}$ are provided to R-SSM, in which $\CE$ is
constructed by connecting two sensors if their road network distance is below
a threshold, $\MV$ stores the geographic positions and learnable embeddings of
sensors, $\TE$ stores the road network distances of edges, and $\MU_{1:T}$
provides the time information (hour-of-day and day-of-week). We impute the missing values for training and exclude them from evaluation. GNF is not used because of GPU memory limitation. Following the settings in \citet{dcrnn}, we train our model on small time windows spanning 2 hours and use a 7:1:2 split for training, validation, and test.

The comparison of mean absolute forecast errors (MAE) is reported in Table \ref{tbl:traffic-mae}. The three forecast horizons correspond to 15, 30, and 60 minutes. We give point predictions by taking the element-wise median of 2K Monte Carlo forecasts. Compared with DCRNN \cite{dcrnn} and GaAN \cite{gaan}, R-SSM delivers comparable short-term forecasts but slightly worse long-term forecasts.

We argue that the results are admissible because:
\begin{enumerate*}[label={(\roman*)},font={\bfseries}]
\item By using MAE loss and scheduled
sampling, the DCRNN and GaAN baselines are trained on
the multi-step objective that they are later evaluated on, making them hard to beat.
\item Some stochastic systems are inherently unpredictable beyond a few steps
due to the process noise, e.g., the toy model in Section \ref{subsec:toy}. In
such case, multi-step MAE may not be a reasonable metric, and probabistic forecasts may be prefered. The average coverage probabilities (CP) of 90\% prediction intervals reported in Table \ref{tbl:traffic-mae} indicate that R-SSM provides good uncertainty estimates.
\item Improving the multi-step prediction ability of deep SSMs is still an
open problem with a few recent attempts \cite{zf-rl,rssm}. We would like to
		explore it in future work.
\end{enumerate*}

%% file: data/toy/metrics.tex
\begin{tabular}{lrrr}  
\toprule
Model                   & LL                    & MSE                              & CP          \\
\midrule
VAR                     & -366                  & 0.679\tiny{ $\pm$.000}           &  0.750\tiny{ $\pm$.000}        \\
VRNN                    & $\geq$-2641          & 0.501\tiny{ $\pm$.003}           &  0.931\tiny{ $\pm$.002}        \\
GNN-AR                  & -94                   & 0.286\tiny{ $\pm$.002}           &  0.806\tiny{ $\pm$.004}        \\
\midrule
R-SSM                  & $\geq$2583           & 0.029\tiny{ $\pm$.001}            &  0.883\tiny{ $\pm$.002}        \\
$+\CL_2^{\text{aux}}$  & $\geq$\textbf{2647}  & \textbf{0.024}\tiny{ $\pm$.001}  &  0.897\tiny{ $\pm$.001}        \\
\bottomrule
\end{tabular}

%% file: data/basketball/metrics.tex
\begin{tabular}{lrrrrr}  
\toprule
Model                  & $\CL_{1000}^{\text{SMC}}$ & ELBO                     & Speed                & Distance               & OOB           \\
\midrule
VRNN                   & ---                       & 2360                     & 0.89                 &  43.78                 & 33.78          \\
MI-VRNN                & ---                       & 2362                     & 0.79                 &  38.92                 & 15.52          \\
\midrule
R-SSM                  & 2459.8\tiny{ $\pm$.3}     & 2372.3\tiny{ $\pm$.8}    & 0.83\tiny{ $\pm$.01}  &  40.75\tiny{ $\pm$.15} &   1.84\tiny{ $\pm$.16} \\
$+\CL_2^{\text{aux}}$  & 2463.3\tiny{ $\pm$.4}     & 2380.2\tiny{ $\pm$.6}    & 0.82\tiny{ $\pm$.01}  &  40.36\tiny{ $\pm$.23} &   2.17\tiny{ $\pm$.09}  \\
\, $+$GNF (4)          & 2483.2\tiny{ $\pm$.3}     & 2381.6\tiny{ $\pm$.4}    & 0.80\tiny{ $\pm$.00}  &  39.37\tiny{ $\pm$.35} &   2.06\tiny{ $\pm$.15}  \\
\, $+$GNF (8)          &\BF{2501.6}\tiny{ $\pm$.2} &\BF{2382.1}\tiny{ $\pm$.4}& 0.79\tiny{ $\pm$.00}  &  39.14\tiny{ $\pm$.29} &   2.12\tiny{ $\pm$.10}  \\
\midrule
Ground Truth           & ---                       & ---                      & 0.77                    &  37.78                   & 2.21          \\
\bottomrule
\end{tabular}

%% file: data/basketball/ball-metrics.tex
\begin{tabular}{lr|r}
\toprule
Model                   & LL      & $\CL_{1000}^{\text{SMC}}$ \\
\midrule
\multicolumn{2}{c|}{\citet{graph-vrnn-cvpr}} & \\
GRNN                    & 2264    &  --- \\
VRNN                    & $>$2750 &  --- \\
GVRNN                   & $>$\BF{2832} &  --- \\
\midrule
R-SSM$+\CL_2^{\text{aux}}$ & $>$2761\tiny{ $\pm$1} & 2805 \tiny{ $\pm$0} \\
$+$GNF (8)              & $>$2783\tiny{ $\pm$1} & 2826 \tiny{ $\pm$0} \\
\bottomrule
\end{tabular}

%% file: data/traffic/table-mae.tex
\begin{tabular}{lrrr}  
\toprule
Model                   & $h = 3$               & $h = 6$               & $h = 12$             \\
\midrule
$\MX_{t-h}$             & 3.97                  & 4.99                  &  6.65                \\
DCRNN                   & 2.77                  & 3.15                  &  \BF{3.60}                \\
GaAN                    & 2.71                  & \BF{3.12}             &  3.64                \\
\midrule
R-SSM                   & \BF{2.67}\tiny{ $\pm$.00}  & 3.14\tiny{ $\pm$.01}  &  3.72\tiny{ $\pm$.02}           \\
\tiny{CP}              & \tiny{0.896 $\pm$.001}  & \tiny{0.891 $\pm$.001}  &  \tiny{0.883 $\pm$.002}           \\
\bottomrule
\end{tabular}

%% file: tex/conclusion.tex
\section{Conclusions}

In this work, we present a deep hierarchical state-space model in which the
state transitions of correlated objects are coordinated by graph neural
networks. To effectively learn the model from observation data, we develop a
structured posterior approximation and propose two auxiliary contrastive
prediction tasks to help the learning. We further introduce the graph
normalizing flow to enhance the expressiveness of the joint transition density
and the posterior approximation. The experiments show that our model can
outperform or match the state-of-the-arts on several time series modeling
tasks. Directions for future work include testing the model on
high-dimensional observations, extending the model to directly learn from
visual data, and including discrete latent variables in the model.

%% file: tex/acknowledgement.tex
\subsubsection*{Acknowledgments}

This work was supported by the National Key Research and
Development Program of China (No. 2018YFB0505000) and the Alibaba-Zhejiang University Joint Institute of Frontier Technologies. Fan Yang would like to thank Qingchen Yu for her helpful feedback on early drafts of this paper.

%% file: tex/appendix.tex
\section{Appendix}

\subsection{MHA implementation}

Each attention head $k \in [K]$ is separately parameterized and operates as follows. For $\forall (j, i) \in \CE$, it produces a message vector $\bm{\beta}_{j \to i}^k$ together with an unnormalized scalar weight $\omega_{j \to i}^k$. Then for $\forall i \in \CV$, it aggregates all messages sent to $i$ in a permutation-invariant manner:
\[
\big\{ \alpha_{p \to i}^k \big\}_{p \in \CN^{-}_i} =
\text{softmax}\left( \big\{ \omega_{p \to i}^k \big\}_{p \in \CN^{-}_i} \right)
\;,\quad
\Va_i^k = {\textstyle \sum_{p \in \CN^{-}_i}}\:{
	\alpha_{p \to i}^k \bm{\beta}_{p \to i}^k
} \;.
\]
Putting all $K$ heads together, it turns out that
\(
\CM_{j \to i} = \{ ( \omega_{j \to i}^k\,,\, \bm{\beta}_{j \to i}^k ) \}_{k=1}^K
\)
and $\CA_i = \{ \Va_i^k \}_{k=1}^K$.
Specifically, each attention head is parameterized in a query-key-value style:
\begin{gather*}
\forall i \in \CV: \quad
\tilde{\Vv}_i = \text{MLP}_v(\Vv_i) \in \BR^{\tilde{d}_v} \; , \quad
\tilde{\Vh}_i = [\Vh_i, \tilde{\Vv}_i, \Vg] \\
\M{Q} = \tilde{\MH} \MW_Q \; , \quad
\M{A} = \tilde{\MH} \MW_A \; , \quad
\M{C} = \tilde{\MH} \MW_C \\
\bm{\beta}_{j \to i} = \Vc_j \;, \quad
\omega_{j \to i} = \V{q}^{\top}_i \V{a}_j  / \sqrt{d_q} + \text{MLP}_e(\Ve_{ji}) \in \BR
\end{gather*}
for $\tilde{d} = d + \tilde{d}_v + d_g$,
$\MW_Q \in \BR^{\tilde{d} \times d_q}$,
$\MW_A \in \BR^{\tilde{d} \times d_q}$, and
$\MW_C \in \BR^{\tilde{d} \times d_c}$.

\subsection{Model details}

In this work, the READOUT function is implemented by passing the concatenation of the outputs of a \emph{mean} aggregator and an element-wise \emph{max} aggregator through a gated activation unit. The transition densities in the generative model are specified to be:
\begin{gather}
f_{\theta}^g\big( \Vz_t^g \big| \Vh_t^g \big) =
\text{Normal} \left(
  \cdot \middle|
  	\bm{\mu}_{\theta}^g \left( \Vh_t^g \right),
  	\bm{\Sigma}_{\theta}^g\left( \Vh_t^g \right)
\right) \; , \\
f_{\theta}^{\star} \big( \MZ_t \big| \MH_t \big) =
\prod_{i=1}^N{
	\text{Normal} \left(
		\Vz_t^{(i)} \middle|
	  	\bm{\mu}_{\theta}^{\star} \left( \Vh_t^{(i)} \right),
	  	\bm{\Sigma}_{\theta}^{\star} \left( \Vh_t^{(i)} \right)
	\right)
} \; ,
\end{gather}
where $\bm{\mu}_{\theta}^g$ and $\bm{\Sigma}_{\theta}^g$ (similarly $\bm{\mu}_{\theta}^{\star}$ and $\bm{\Sigma}_{\theta}^{\star}$) are 3-layer MLPs that
share their first layer. $\bm{\Sigma}_{\theta}^g$ and $\bm{\Sigma}_{\theta}^{\star}$ output diagonal covariance matrices using the softplus activation. The proposal densities $r_{\phi}^g$ and $r_{\phi}^{\star}$ are specified in a similar way. Then GNFs can be stacked on top of $f_{\theta}^{\star}$ and $r_{\phi}^{\star}$ to make them more expressive.

\subsection{Experiments}

We use the Adam optimizer with an initial learning rate of 0.001 and a gradient clipping of 1.0 for all experiments. The learning rate was annealed according to a linear cosine decay. We set $\beta_1 = \beta_2 = 1.0$ for the auxiliary losses in all experiments.

\begin{figure}
\center
\includegraphics[width=\linewidth]{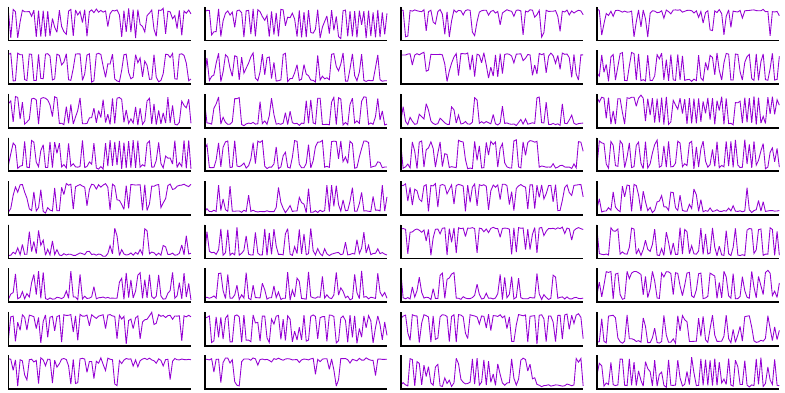}
\caption{An example from the toy dataset ($N = 36$, $T = 80$).}\label{fig:toy-example}
\end{figure}

\paragraph{Synthetic toy dataset.}
A typical example in the dataset is visualized in Figure \ref{fig:toy-example}. The architectures of the models are specified as follows.
\begin{enumerate}[label={(\alph*)}]
\item VRNN: Using 128-dimensional latent variables and a two-layer, 512-unit GRU.
\item GNN-AR: Using a two-layer GNN and an one-layer, 128-unit GRU shared by all nodes. 
\item R-SSM: We let $d_g = d_z = 8$. All RNNs are specified to be two-layer, 32-unit LSTMs. All MLPs use 64 hidden units. The generative model and the proposal both use a 4-head MHA layer. 4 SMC samples and a batch size of 16 are used in training.
\end{enumerate}

\begin{figure}
  \centering
  \includegraphics[width=.225\linewidth]{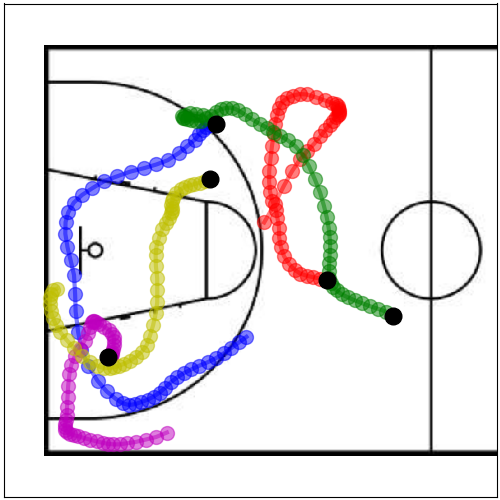}
  \:
  \includegraphics[width=.225\linewidth]{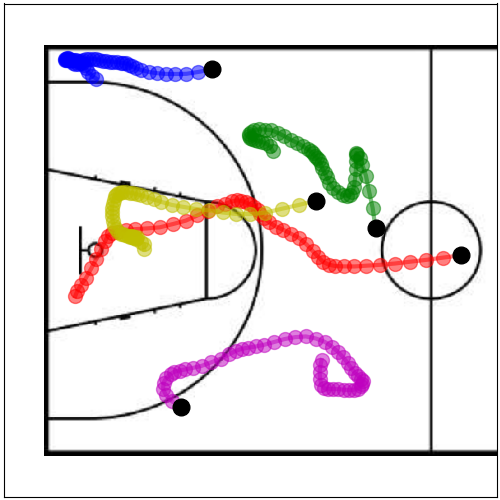}
  \:
  \includegraphics[width=.225\linewidth]{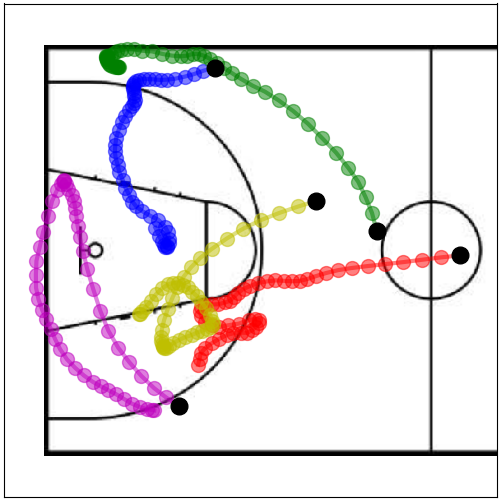}
  \:
  \includegraphics[width=.225\linewidth]{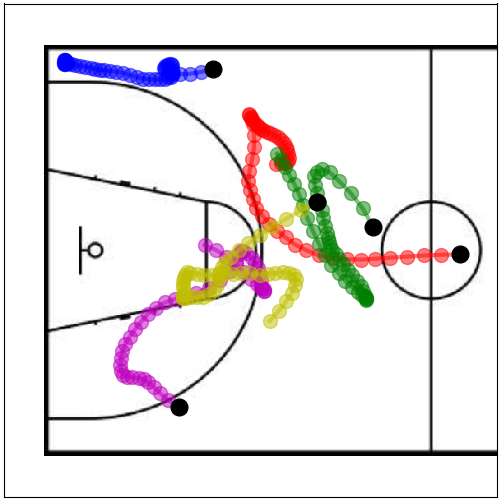}
  \\[\baselineskip]
  \includegraphics[width=.225\linewidth]{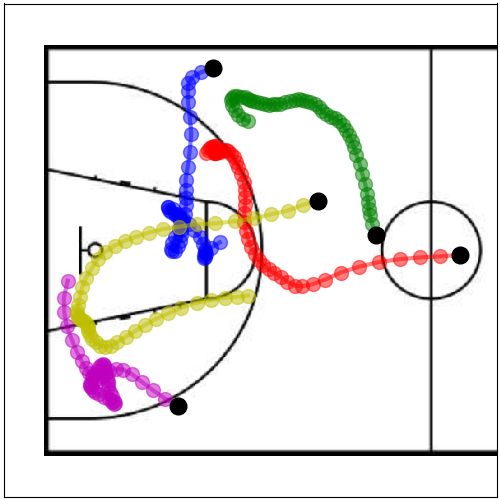}
  \:
  \includegraphics[width=.225\linewidth]{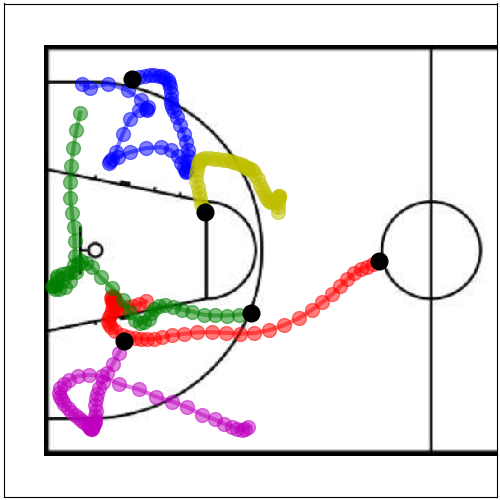}
  \:
  \includegraphics[width=.225\linewidth]{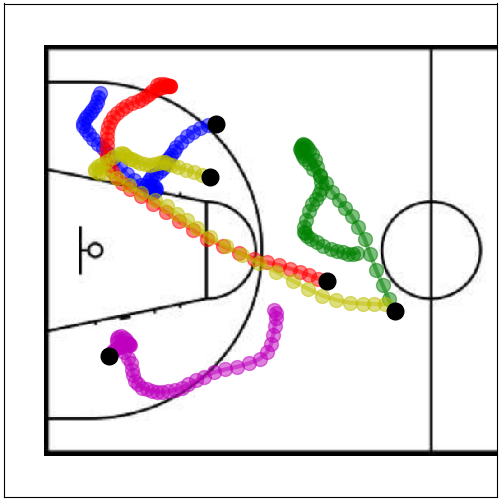}
  \:
  \includegraphics[width=.225\linewidth]{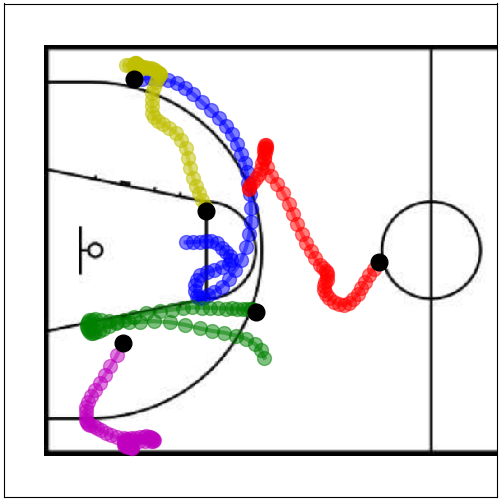}
  \caption{Selected rollouts from the trained model. Black dots represent the starting points.}
  \label{fig:bball-rollouts}
\end{figure}

\paragraph{Basketball player movement.}
We let $d_g = d_z = 32$. All RNNs are specified to be two-layer, 64-unit LSTMs and all MLPs use 256 hidden units. The generative model uses one 8-head MHA layer and the proposal uses two 8-head MHA layers. Each GNF uses an additional MHA layer shared by the functions $s(\cdot)$ and $t(\cdot)$. 4 SMC samples and a batch size of 64 are used in training. Eight selected rollouts from the trained model are visualized in Figure \ref{fig:bball-rollouts}.

\paragraph{Road traffic.}
We let $d_g = d_z = 8$. A 32-dimensional embedding for each sensor is jointly learned as a part of the vertex attribute. All RNNs are specified to be two-layer, 32-unit LSTMs and all MLPs use 64 hidden units. The generative model and the proposal both use two 8-head MHA layers. 3 SMC samples and a batch size of 16 are used in training.

\newpage

\subsection{Training}\label{subsec:app-train}

We optimize the VSMC bound estimated by the following SMC algorithm:

\begin{algorithm}
	\caption{Estimate the VSMC bound $\CL_K^{\text{SMC}}$}
    \label{alg:vsmc}
\begin{algorithmic}
    \STATE {\bfseries Input:} graph $\CG$, observations $\MX_{1:T}$, exogenous inputs $\MU_{1:T}$
	\STATE {\bfseries Require:} generative model $\{ f_{\theta}^g, f_{\theta}^{\star}, g_{\theta} \}$,
	proposal $\{ r_{\phi}^g, r_{\phi}^{\star} \}$, number of particles $K$
	\STATE

	\FOR{$k = 1 \ldots K$}
	\STATE Simulate $\Vz_{1, k}^g \sim r_{\phi}^g (\cdot | \CG, \MX_1, \MU_1 )$
	\STATE Simulate $\MZ_{1, k} \sim r_{\phi}^{\star} (\cdot | \Vz_{1, k}^g, \CG, \MX_1, \MU_1)$ 
	\STATE Set \(
	w_1^k = \frac{
		f_{\theta}^g(\Vz_{1, k}^g)
		f_{\theta}^{\star}(\MZ_{1, k} | \Vz_{1, k}^g,\, \CG,\, \MU_1)
		\prod_{i=1}^{N}{ g_{\theta}( \Vx_1^{(i)} | \Vz_{1, k}^{(i)},\, \Vz_{1, k}^g,\,\ldots) }
	}{
		r_{\phi}^g(\Vz_{1, k}^g | \ldots)
		r_{\phi}^{\star}(\MZ_{1, k} | \ldots)
	}
\)
	\ENDFOR

	\STATE Initialize $\hat{\CL}^{\text{SMC}}_K = \log{ \sum_{k=1}^{K}{\nicefrac{w_1^k}{K}} }$
	\STATE

	\FOR{$t = 2 \ldots T$}
	\STATE $\{ \Vz_{<t, k}^g, \MZ_{<t, k} \}_{k=1}^K = \text{RESAMPLE}(\{ \Vz_{<t, k}^g, \MZ_{<t, k}, w_{t-1}^k \}_{k=1}^K )$
	\FOR{$k = 1 \ldots K$}
	\STATE Simulate $\Vz_{t, k}^g \sim r_{\phi}^g (\cdot | \Vz_{<t, k}^g, \MZ_{<t, k}, \CG, \MX_{\leq t}, \MU_{\leq t} )$
	\STATE Simulate $\MZ_{t, k} \sim r_{\phi}^{\star} (\cdot | \Vz_{\leq t, k}^g, \MZ_{<t, k}, \CG, \MX_{\leq t}, \MU_{\leq t})$ 
	\STATE Set \(
	w_t^k = \frac{
		f_{\theta}^g(\Vz_{t, k}^g | \ldots)
		f_{\theta}^{\star}(\MZ_{t, k} | \Vz_{\leq t, k}^g,\, \MZ_{<t, k},\, \ldots)
		\prod_{i=1}^{N}{ g_{\theta}( \Vx_t^{(i)} | \Vz_{t, k}^{(i)},\, \Vz_{\leq t, k}^g,\, \MZ_{<t, k},\, \ldots) }
	}{
		r_{\phi}^g(\Vz_{t, k}^g | \ldots)
		r_{\phi}^{\star}(\MZ_{t, k} | \ldots)
	}
\)
	\STATE Set $\Vz_{\leq t, k}^g = (\Vz_{<t, k}^g, \Vz_{t, k}^g )$,
	       $\MZ_{\leq t, k} = (\MZ_{<t, k}, \MZ_{t, k})$
	\ENDFOR
	\STATE Update $\hat{\CL}^{\text{SMC}}_K = \hat{\CL}^{\text{SMC}}_K + \log{ \sum_{k=1}^{K}{\nicefrac{w_t^k}{K}} }$
	\ENDFOR
	\STATE
	\STATE {\bfseries Output:} $\hat{\CL}^{\text{SMC}}_K$
\end{algorithmic}
\end{algorithm}

In our model, dependencies on $\Vz_{<t, k}^g$ and $\MZ_{<t, k}$ are provided
through the compact RNN states $\Vh_{t, k}^g$ and $\MH_{t, k}$. When GNFs are
used in $f_{\theta}^{\star}$ and $r_{\phi}^{\star}$, density calculation and
backpropagation are automatically handled by the TensorFlow Probability
library.

To estimate the auxiliary objectives
$\CL_1^{\text{aux}}$ and $\CL_2^{\text{aux}}$,
we reuse the resampled unweighted particles
$\{ \{ \Vz_{1:t, k}^g, \MZ_{1:t, k} \}_{k=1}^K \}_{t=1}^{T-1}$
generated by Algorithm \ref{alg:vsmc} to form Monte Carlo estimations for them:
\[
\hat{\CL}_1^{\text{aux}} = \sum_{t=1}^{T-1} \sum_{i=1}^N \frac{1}{K}
  \log \frac{
    \lambda_{\psi, 1}( \hat{\Vz}_{t, k}^{(i)}, \Vc_t^{(i)} )
  }{
    \sum_{ \Vc \in \Omega_{t,i} }{
       \lambda_{\psi, 1}( \hat{\Vz}_{t, k}^{(i)}, \Vc ) 
    }
  }
, \;
\hat{\CL}_2^{\text{aux}} = \sum_{t=1}^{T-1} \sum_{i=1}^N \frac{1}{K}
  \log \frac{
    \lambda_{\psi, 2}( \hat{\Vh}_{t, k}^{(i)}, \Vc_t^{(i)} )
  }{
    \sum_{ \Vc \in \Omega_{t,i} }{
       \lambda_{\psi, 2}( \hat{\Vh}_{t, k}^{(i)}, \Vc ) 
    }
  }
\]
where
$\hat{\Vz}_{t, k}^{(i)} = [\Vz_{t, k}^g, \Vz_{t, k}^{(i)}]$, 
$\hat{\Vh}_{t, k}^{(i)} = \text{MLP}_{\psi}( \sum_{ j \neq i \, \wedge \, j \in \CN_i^{-} }{ \Vh_{t, k}^{(j)} } )$, and $\Omega_{t,i}$ is a set that contains $\Vc_t^{(i)}$ and some negative samples selected from the future summaries of other vertices within the minibatch.

\newpage

\subsection{Graph Normalizing Flow}\label{subsec:app-gnf}

\begin{figure}[H]
\center
\includegraphics[width=\linewidth]{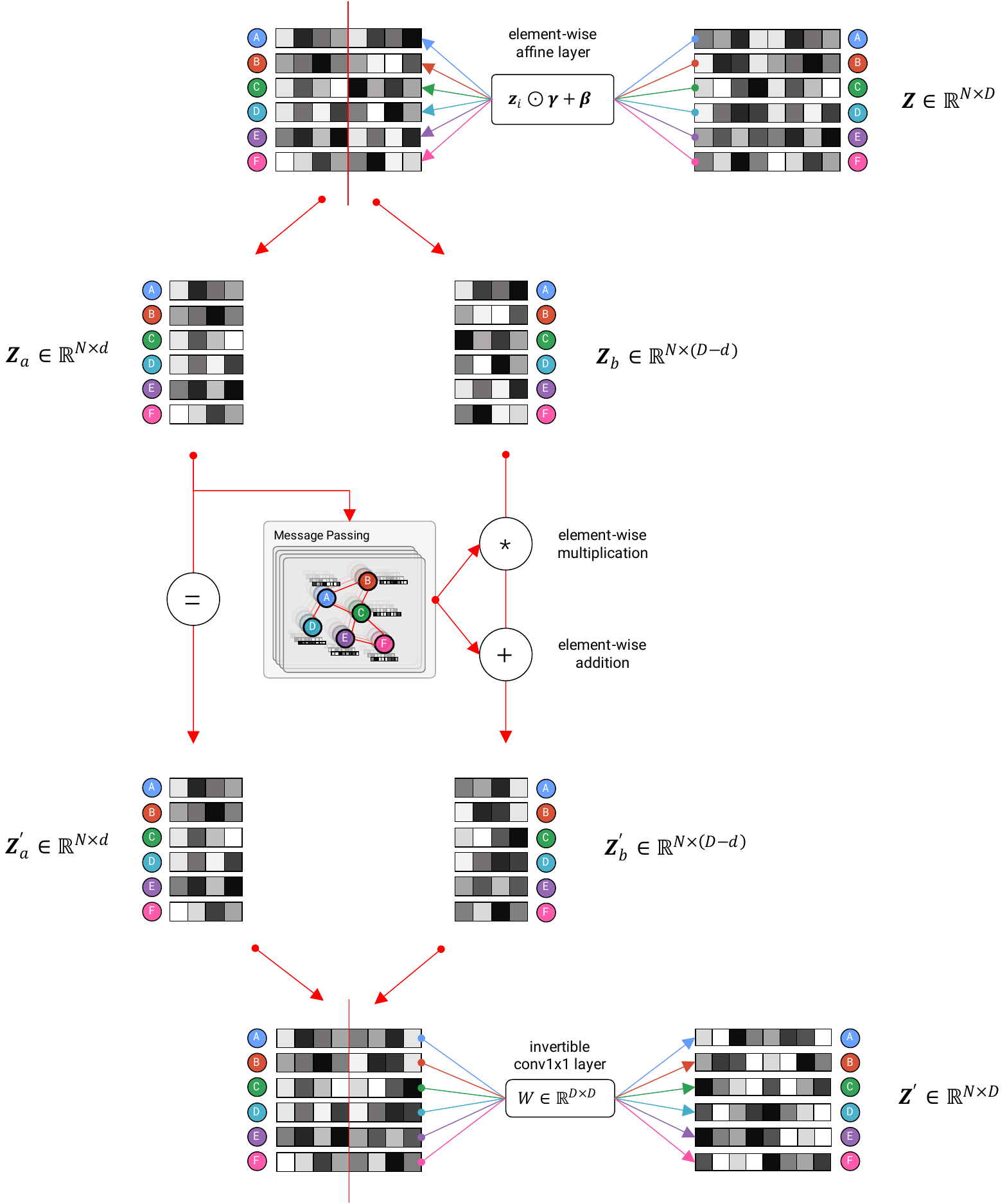}
\caption{Visual illustration of a GNF. Multiple GNFs can be stacked together to
	achieve more expressive transformation.}\label{fig:gnf}
\end{figure}

The element-wise affine layer is proposed by \citet{glow} for normalizing the
activations. Its parameters $\bm{\gamma} \in \BR^D$ and $\bm{\beta} \in \BR^D$
are initialized such that the per-channel activations have roughly zero mean
and unit variance at the beginning of training. The invertible linear
transformation $\MW \in \BR^{D \times D}$ is parameterized using a QR
decomposition \cite{emerging}.